\documentclass[conference]{./sty/IEEEtran}

\usepackage[letterpaper, margin=0.75in]{geometry}
\usepackage{epsfig}
\usepackage{amsmath}
\usepackage{amssymb}
\usepackage{float}
\usepackage{afterpage}
\usepackage[breaklinks=true,colorlinks,bookmarks=false]{hyperref}

\newcommand{\etal}{\textit{et al. }}

\author{
  \IEEEauthorblockN{Thomio Watanabe}
  \IEEEauthorblockA{University of Sao Paulo\\
  {\tt\small thomio.watanabe@usp.br}}
  \and
  \IEEEauthorblockN{Denis Wolf}
  \IEEEauthorblockA{University of Sao Paulo\\
  {\tt\small denis@icmc.usp.br}}
}

\title{Verisimilar Percept Sequences Tests for\\Autonomous Driving Intelligent Agent Assessment}

\begin{document}

\newgeometry{top=1in,bottom=0.75in,right=0.75in,left=0.75in}
\maketitle


\begin{abstract}

The autonomous car technology promises to replace human drivers with safer driving systems.
But although autonomous cars can become safer than human drivers this is a long process that is going to be refined over time.
Before these vehicles are deployed on urban roads a minimum safety level must be assured.
Since the autonomous car technology is still under development there is no standard methodology to evaluate such systems.
It is important to completely understand the technology that is being developed to design efficient means to evaluate it.
In this paper we assume safety-critical systems reliability as a safety measure.
We model an autonomous road vehicle as an intelligent agent and we approach its evaluation from an artificial intelligence perspective.
Our focus is the evaluation of perception and decision making systems and also to propose a systematic method to evaluate their integration in the vehicle.
We identify critical aspects of the data dependency from the artificial intelligence state of the art models and we also propose procedures to reproduce them.

\end{abstract}

\section{Introduction}
To ensure quality and safety the automotive industry thoroughly tests motor vehicles.
This is a usual procedure carried by all major manufacturers.
Most tests focus on physical attacks to evaluate equipment durability, electrical and radio frequency emission and immunity.
These tests are defined in international standards and company procedures.
With the advent of new technologies to automate motor vehicles these tests must also consider higher level functionalities. 
The automation technology aims to replace human drivers and therefore tests that contemplate machine intelligence must be required.
Given its importance this technology should also be considered a safety-critical system.

The ultimate goal of autonomous road vehicles is to provide a safe and reliable transportation.
But safety is a subjective concept that is hard to assess numerically and systematically \cite{stellet2015testing}.
One way to assess it is through events that demonstrates the lack of safety like the rate of crashes, injuries and deaths per traveled distance.
The biggest problem of these metrics is that they can only be assessed after deploying vehicles in urban roads as an after action review.
And deploying vehicles on urban roads for a late safety assessment is irresponsible.

An informal metric that is being used by car manufacturers is the ``disengagement rate'' which is the number of times a driver must regain control over the vehicle.
This ``metric'' is extremely vague because a disengagement can be triggered by any event at any time.
It is also expected that a fully automated vehicle never disengage and the fact that a vehicle does not disengage does not mean it is safe.

In the machine learning context Varshney \cite{varshney2016engineering} defines safety as the minimization of risk and uncertainty of harmful events.
He also presents some strategies for achieving safety but does not mention evaluation methods.
According to this definition safety is a state of local minimum risk.
Although it seems a good idea, it might be inappropriate to always drive at the minimum risk state.
Taking risks is part of the driving experience and there is a compromise among different driving aspects like safety, speed and comfort.
An evaluation method should certify harmful events never happen independently of the vehicle safety level.

From a research perspective most autonomous cars evaluation methodologies use base models to assess performance on specific tasks \cite{huang2014task, roesener2017safe}.
Although this evaluation is interesting for development it is biased towards the base model and do not present a fair evaluation between acceptable solutions.
We assess safety by evaluating safety-critical system performance on driving tasks described through percept sequences.
Where the base model is replaced by environment constraints that must be respected during the task execution.

Although reliability and safety are different concepts they are directly correlated.
If a vehicle safety-critical system fails there will be a great risk of death or serious injury to people.
Therefore autonomous vehicles must guarantee safety-critical system faults are as low as reasonably practicable.



We focus is the evaluation of the intelligent agent designed to automate road vehicles, comprising the perception and decision-making systems, which are the core problem for autonomous road vehicles \cite{okumura2016challenges}.
We also model the problem to propose a systematic approach for a thorough evaluation of the system integration.
The proposed intelligent agent evaluation can be made for different groups of sensors and tasks (automation level) and it is independent of the model/algorithm implementation.




\afterpage{\aftergroup\restoregeometry}

\section{Related Work}

Given the importance of the subject there are few papers that assess autonomous driving systems performance.
Numerically evaluate such systems is a hard task because it involves unknown variables, a relative definition of the best behavior and also a clear division of right and wrong decisions.
For instance, some machine learning methods usually presented in perception systems are still considered black boxes \cite{castelvecchi2016can, olah2018the}.
The local culture background is an important aspect to be considered when defining what is the best solution for a problem \cite{bruno2017framework}.
Also, there are moral problems for autonomous road vehicles that may never have a right answer, problems like ethical or moral dilemmas \cite{malle2015sacrifice, bonnefon2016social, lin2016ethics}.

The international standard ISO 26262 aims the functional safety in road vehicles and provides guidelines for the vehicle life cycle.
It presents instructions for hazard analysis and risk assessment but it does not define how to identify and deal with hazardous situations.
It does not present evaluation guidelines and, for being too generic, it leaves space for subjective and discretion decisions.
Also given its release date, 2011, it does not consider the current developments in Artificial Intelligence (AI) that are being applied in Advanced Driver Assistance Systems and Autonomous Driving.
This standard was used by Johansson \etal \cite{johansson2016need} to identify safety requirements for environment perception and it was also used by Khastgir \etal \cite{khastgir2017introducing} to rate motor vehicles hazardous situations in real-time.

In \cite{khastgir2015identifying}, Khastgir \etal addresses the lack of a standardized process to validate intelligent automotive systems.
It also compares existing methods and presents a driving simulator as a test platform but it does not develop testing methodologies and it does not address real data problems from perception systems.

Huang \etal \cite{huang2016autonomous} presents a review of methods that can be used to test individual subsystems and their integration.
To test algorithms, a common approach is based in individual task performance assessment where the execution of each task is compared to a correct model.

Most papers in the area focus in a developer point of view of testing.
They specify an exact behavior that must be followed and then measure how much proposed solutions deviates from it \cite{huang2014task, roesener2017safe}.
Although this is interesting for developers the ``correct'' behavior model is subjective and produce biased assessments.
These models may also be private data protected by the intellectual property of vehicle manufacturers.
For an evaluation point of view this approach is too much restrictive.
There are two main concerns regarding this method.
First, it might be impractical to create a model for every single possible situation; and second, one situation may have several correct actions (and no best action).
For instance if a frontal obstacle suddenly stops, changing lanes or also stopping are both acceptable actions.

Computer simulation is becoming the main platform to test and validate autonomous road vehicles \cite{stellet2015testing, huang2016autonomous, gomez2018driving}.
There are several reasons to support this decision but the perception system evaluation is far from ideal.
Urban roads are extremely rich and dynamic environments and due to the generalization error a perception system evaluation cannot be trusted if the testing examples do not represent the actual vehicle deployment region.
In general a simulator must create realistic environments and sensors models but there is no metric to evaluate how realistic they are.

\section{Evaluation Model}
\label{sec:eval_model}

\begin{figure}
\begin{center}
	\includegraphics[scale=.8]{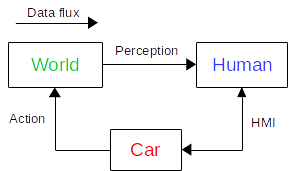}
	\caption{The usual car model is controlled by a human that is responsible to perceive the surroundings and make decisions.}
\label{fig:car_world}
\end{center}
\end{figure}

Lets consider the Fig. \ref{fig:car_world} a simplified version of the current car technology.
In this model the human driver is the only responsible to perceive the surroundings.
Also, the driver is responsible to make decisions and to control the vehicle accordingly.

\begin{figure}
\begin{center}
	\includegraphics[scale=.8]{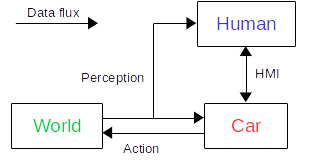}
	\caption{An autonomous car is able to perceive the world. The human can still take control of the car but it must be able to work without human intervention.}
\label{fig:robot_world}
\end{center}
\end{figure}

The automation technology that is being developed aims to replace the human driver.
Therefore the vehicle must perform to some extent the perception and decision-making tasks that were the driver responsibility.
A general view of these interactions is depicted at Fig. \ref{fig:robot_world}.

The main difference between these models is that the autonomous car must perceive its environment and make decisions.
From this top-down view it is a simple modification but it is a hard task to accomplish.

The Generic Autonomous Vehicle System Architecture divide an autonomous car software in three main layers: perception, decision-making and action \cite{huang2016autonomous}.
This general division is interesting for assessment solutions because it does not impose technological prerequisites.

Since the action layer is developed according to the control engineering theory which is an established theory with decades of use we focus our analysis in the perception and decision-making layers.

\subsection{Perception}

In the perception task, to replace the human sight an autonomous vehicle makes uses of several sensors.
The most common sensors are video cameras, radar, LIDAR and ultrasonic sensors. 
However the raw data from most sensors do not present the meaningful information required by a decision-making system.
In this work the perception system comprises the software required to transform raw sensor data in the meaningful information required by the decision-making system.


Initial researches on autonomous cars date from the 80s \cite{michon1985critical, pomerleau1989alvinn}, but only recently that they are becoming a reality.
Computational power and the perception system were great problems for autonomous cars, limiting the automation.
Extracting meaningful data from the sensors was a real challenge that could only be overcome by recent advancements in artificial intelligence \cite{lecun2015deep}.


It is possible to separate tests in two groups: code implementation and model/algorithm.
There are several software engineering techniques to test code implementation, usual examples are unit tests and integration tests.
In the machine learning community the usual way to evaluate models and algorithms are through datasets.
Although datasets are independent of the model implementation, datasets are task specific (classification, segmentation, detection) that are subtasks of a complete perception system.

Recent papers present evidences that machine learning models do not have a task optimality \cite{zoph2017learning, liu2017hierarchical, real2018regularized}. 
In other words a specific model may present different performances on different datasets describing the same task.
A model that has a good performance classifying dogs and cats may have a bad performance classifying cars and pedestrians.

This data dependence was previously presented in different ways.
A common problem is the generalization error which measures the ability of a supervised learning model detect patterns not presented in the training set. 
The data imbalance is also a common problem that hinders model training, inducing the model to neglect rare patterns. 
Therefore a dataset evaluation is not only task dependent but also data dependent.

The generalization error is usually estimated as the difference between the training and testing sets accuracy.
This estimate is very poor once it depends on the available data.
In other words, it is not possible to assess how a model will perceive an example that was not present in the dataset.
This aspect brings concerning issues when using datasets.
As a general rule of thumb a dataset must exactly represent the patterns intended to be found, including different weather conditions.
Since different cities present different landscapes, urban architecture, car designs, people clothes, different animals...
For the autonomous vehicles context the safer approach is creating a dataset for each region these vehicles are intended to be deployed.

Although datasets evaluate the generalization error (to a certain degree) and also the class imbalance, there are other problems that arise with the current model-task-data evaluation with datasets.
\begin{enumerate}
  \item They focus on subtasks of a bigger problem.
  \item The generalization error estimate strongly depends on available data.
  \item Training and testing sets are equally balanced.
  \item Testing examples have a high quality.
  \item Testing examples are always reliable.
  \item Testing examples are always available.
\end{enumerate}

In the autonomous vehicles context these data issues have been overlooked.
In this work we try to enlighten this problem pointing out specific issues that may arise during the data acquisition, processing and transmission to perception modules.
We also propose a more realistic evaluation of autonomous driving perception systems based in percept sequences.


\subsection{Decision-making}
The decision-making process is studied in several fields of the human knowledge including psychology, mathematics and artificial intelligence.
Since we approach the problem from an AI perspective we can model the decision-making system according to the decision theory which combines probabilities with preferences expressed by utilities \cite{russell2009artificial}.
The decision theory studies the reasoning underlying an agent's choices.
The probability function describes the chances to change states through actions and the utility function numerically define state preferences.

An agent state is determined by the set of information it gathers about the environment through its sensors.
A state may be considered a condition where the agent finds itself.
The agent actions are able to change its state but these actions are only executed if they are expected to improve the agent's condition over time.
A utility function defines which conditions are considered good or bad, describing preferences over conditions.
Preferences are fundamentally relative between individuals or social groups and incorporate moral principles/social norms.
Therefore, the decisions are based in preferences that are influenced by a cultural background \cite{bruno2017framework}.

It is also interesting to note that some studies point out the humans do not usually make rational decisions and are ``predictably irrational'' \cite{russell2009artificial}.
Irrational decisions on urban roads are easily spotted on everyday traffic.
Human preferences towards some irrational decisions may lead to future conflicts with intelligent vehicles.


The utility function is very flexible and can contemplate many different aspects of the driving experience as comfort, safety and speed.
In case safety is a variable there must be an approach to define the risk of each possible state the vehicle can assume, similar as presented by Khastgir \etal \cite{khastgir2017introducing}.
Many different utility functions may be defined and all of them may be considered safe and reliable.
However when a utility function is defined it specifies a behavior that can be used as a base model.
Enforcing a base model as the right behavior prevents modification in the utility function and does not give space for adjustments, innovations or improvements.

It is a common sense that decision-making systems must follow moral principles \cite{van2006ethics, van2007ethics,crnkovic2012robots}.
But how to make rational decisions that are against moral principles?
Given the condition an autonomous car needs to kill a pedestrian or the vehicle passenger, if a human life is invaluable which one should die?
In this case not choosing is also a decision that will kill the pedestrian.
Should it kill the pedestrian that has nothing to do with the vehicle previous actions or should it kill the passenger that bought an autonomous car because it is safe?
If a vehicle can make a utilitarian decision how should it assess a person's life value ?
Obviously these situations must be prevented at all costs but in case it is not possible, car manufacturers are not the ones supposed to make them.
The best option is to get these decisions regulated by local state agencies.
Also, the robot moral principles will be likely different from the human moral principles \cite{malle2015sacrifice}.
For most social groups the self-defense premise allows humans to kill their aggressors but this principle will likely not be applied to robots.

Moral dilemmas happen when it is not possible to make a decision without breaking a social norm.
But there are situations where we must break a regulatory law to comply with social norms and also situations that we must decide which regulatory law to break.
These situations present the interesting design problem of defining decision priorities.

There is also the problem of fault and liability for traffic accidents.
Initially it may seem like a simple problem to identify who made wrong decisions that lead to an accident but there are more variables involved.
Since there are different automation levels a human driver and an autonomous vehicle may share decisions and therefore responsibilities \cite{geistfeld2017roadmap, awad2018blaming}.
Anyway, legal liability and decision-making design are out of the scope of this work.

A methodology to evaluate decision-making systems must be independent of the technology and flexible enough to contemplate social groups preferences.
But the biggest problem to evaluate decision-making systems resides in the fact that it may exist several right decisions.
In such conditions it is typical to look for errors/faults, or specifically, for events that represent the lack of safety or regulatory compliance.
This approach conforms with constitutional principles such as ``Everything which is not forbidden is allowed''.

\section{Evaluation Pipeline}

To evaluate an autonomous road vehicle we propose the use of \textbf{percept sequences} which were defined by Russel and Norvig \cite{russell2009artificial} as \textit{``the complete history of everything the agent has ever perceived''}.
But rather than a single and long percept sequence the evaluation should consist of several sequences.
Each percept sequence is a stream of sensors data with timestamps.

The perception system would be evaluated by a dataset created with real sensors data.
In this specific evaluation it is not a requirement that a percept sequence represents task environments.
The most important aspect of this evaluation is to assess how the perception system deals with the problems related to the input data from real world sensors. 

Differently from the perception system, the decision-making system must be evaluated over task environments.
Task environments are the ultimate goal of rational agents i.e. they are the tasks that a rational agent was built to accomplish.
The decision-making evaluation should be made in a simulator.
And since the perception system is previously evaluated with real sensors data, realistic sensor models and environments are not strong requirements.
What matters is the complexity of the relationship between the agent behavior and the environment.

A concern when creating percept sequences is related to the number of sensors used by a vehicle.
In general, a system reliability scales with the number of sensors.
Therefore, for a fair evaluation the number of sensor and their disposition on vehicles should be represented in the dataset.




Our main focus is the evaluation of perception and decision-making systems.
To contextualize we propose a longer evaluation pipeline including the following steps:

\begin{enumerate}
  \item individual perception subtasks
  \item data verisimilitude test
  \item decision-making simulation
  \item outdoor controlled test
  \item autonomous test and deployment
\end{enumerate}

\subsection{Individual Perception Subtasks}
This step evaluates the performance of perception subtasks.
It should be done with datasets as it is the current usual procedure.
Datasets are very important for the machine learning and robotics community.
They provide the means for a fair evaluation between different models.
Several urban roads datasets were created to leverage developments on autonomous road vehicles \cite{Geiger2012CVPR, Cordts2016Cityscapes, ros2016synthia, huang2018apolloscape}.
Since supervised learning algorithms performance scales with the amount of data, the development of new datasets is always beneficial. 

The usual datasets fulfill their purpose to evaluate models/algorithms performance, establishing a procedure to compare them.
They are also important to train supervised learning models, although this aspect is not directly evaluated by datasets.
Datasets present high quality data handpicked and labeled by humans and they fail to model problems that arise from real world applications.


\subsection{Data Verisimilitude Test}
The perception task can be subdivided in several subtasks like obstacles detection, classification and tracking, traffic signs detection and road detection.
This ``divide and conquer'' approach is a valid simplification to solve complex problems.
In autonomous vehicles these subtasks are later integrated in a module usually called fusion module.

Although the subtasks evaluation are already established not much attention was given to how the integration module is evaluated. 
Since the main idea of using several and different sensors is to provide more reliable results we propose to evaluate how the fusion module performs facing real data and problems that can emerge from its use.

With different sensors a fusion module is able to cover a wider operational range providing a super human ability to autonomous vehicle.
An autonomous vehicle may have a 360-degree visual field, it may recognize objects at distant ranges or even in the dark.
Besides the usage of different sensors, the fusion module also make use of sequential data to estimate future states.
Therefore the usage of percept sequences to evaluate fusion modules seems to be logic.

The data verisimilitude test is a dataset consisting of several percept sequences.
Its evaluation metrics are the same used in the individual subtasks and its main intention is to evaluate how the fusion module performs facing the data problems mentioned in the section \ref{sec:eval_model}.
In other words, the dataset must contain data representing problems that are usually found during an autonomous road vehicle deployment.
These problems were separated in 5 groups: data estimation, data absence, data flux interruption, data corruption and data attacks.


\textbf{Data estimation}.
Since an urban road is a partially observable environment an intelligent agent must be able to estimate variables.
These variables can be partially observable in time or by different sensors.
Some patterns may be present on parts of percept sequences or some patterns may appear in only one of the sensors.
As examples there are objects with total occlusion and objects in dark regions detected only by a LIDAR.
Future estimates are extremely important for decision-making once they predict trajectories and can anticipate collisions.
Since percept sequeces are streams of sensor data, future states are part of the dataset and can be used to assess these estimates.


\textbf{Data absence} is represented by data patterns that are absent on the training set.
Since it is impossible to create a dataset with all objects that may appear in the road, its main objective is to present a more realistic estimate of the generalization error. 
This may be represented by different or unusual animals, buildings or vehicles. 
Rare and unusual data examples could be moved from the training to the testing set to reproduce this issue.
This procedure would cause a data set imbalance between the training and testing sets.
Note this is different from class imbalance.

\textbf{Data flux interruption} represents a general sensor malfunction or a sensor communication channel loss.
There are several reasons why a data flux interruption may occur.
For instance a sensor may be broken, its communication channel or power line may be cut, or a sensor may be blocked by external objects compromising its normal operation (LIDAR stuck).
This could be reproduced by inconsistent sensors data, like only zeros/ones.

\textbf{Data corruption} is any data modification caused by signal interference.
Many variables may cause data interference but we consider only random and unintentional interference.
An interference may occur directly on the digital data or indirectly on the sensor analog input.
It may happen during data acquisition, processing or transmission.
As interference causes we cite natural interactions with physical quantities like highly illuminated objects, radiation-absorbent material, moisture, mud, dirt, sea spray, and etc.
It may happen because weather conditions, by a sensor malfunction or when a sensor is out of its operational range.
It also may occur by radio frequency interference and electrical discharges originated externally or from other devices installed on vehicle.




\textbf{Data attacks} (sensor or remote attacks) are input data manipulated by external agents with the intention to cause errors.
Some papers \cite{petit2015remote,shin2017illusion} partially address the issue showing it is possible to remotely manipulate cameras and LIDAR input data.
However, they do not evaluate the perception algorithms that process the input data.
For instance Petit \etal \cite{petit2015remote} generates attacks that degrade the sensors data quality while Shin \etal \cite{shin2017illusion} is able to induce fake dots and saturate a LIDAR sensor.
These works directly attack sensors corrupting input data but it is also possible to do it indirectly, generating data patterns that can fool perception algorithms.

In computer vision these modified images are denominated adversarial examples and the manipulated information is denominated perturbations \cite{szegedy2013intriguing, goodfellow2015explaining}.
These attacks are carried by adding perturbations in data patterns that are detected by a perception system.
Several recent researches provide evidences that real world perturbations can be designed to deceive perception models \cite{kurakin2016adversarial, evtimov2017robust, eykholt2017note}.
This is one of the most critical problems since current solutions are extremely limited and do not present definitive solutions \cite{athalye2018obfuscated, kannan2018adversarial}.



\subsection{Decision Making Simulation}

For a thorough assessment of autonomous vehicles performance the decision-making system must be considered.
The perception system will never be perfect but the decision-making system must able to define countermeasures to partial or defective information, as noted by Okumura \etal \cite{okumura2016challenges}.
For instance, a solution could be implemented to check if a sensor is working under its operational range and otherwise the autonomous vehicle should perform a controlled stop.


The decision-making system should also be evaluated with percept sequences.
But differently from the perception system the percept sequences must represent task environments.
This is an approach similar to Huang \etal \cite{huang2014task} however the model based evaluation should be replaced by a constraint based evaluation, where each task must be executed within time-space-interaction restrictions (or environment state).
This change would allow a fair evaluation of solutions with different base models.
For instance, a car that parks itself within 1 minute in the limited parking area without hitting anything must be consider a valid solution.

A task evaluation satisfies the Russel and Norvig statement \cite{russell2009artificial}: \textit{``As a general rule, it is better to design performance measures according to what one actually wants in the environment, rather than according to how one thinks the agent should behave''}.

This decision-making evaluation is not able to rank solutions.
Its main purpose is to reprove solutions that are not able to comply with the given task.
Therefore it is a binary decision-making evaluation where the vehicle is free to take any action as long as it do not cause any harm and they respect the traffic laws.
It is a binary evaluation because the vehicle automatically fails if it does not comply with the rules and it actually assess the lack of safety.

It is interesting to note that a real urban road is an extremely dynamic environment and a real time measurement of all relevant variables is impractical.
Given the complexity of the task the only feasible solution is to evaluate the decision-making system through computer simulations.
But as Russel and Norvig \cite{russell2009artificial} states: \textit{``In fact, what matters is not the distinction between real and artificial environments, but the complexity of the relationship among the behavior of the agent, the percept sequence generated by the environment, and the performance measure''}.

Since perception systems are highly data dependent certainly Russel and Norvig were addressing high level functionalities such as decision-making.
And when the perception system is previously tested with real sensor data the distinction between the real and artificial environment becomes irrelevant for a decision-making system evaluation.
What is important to attend is the performance level of the simulated sensors that must be similar to that observed in the data verisimilitude test.


In summary a simulation is the best choice for decision-making assessment because it can evaluate the lack of safety without causing real damage. 
It is able to manage the high number of agents and the complex interaction among them.
And also it gives a concrete sense of the consequences of each action.

\subsection{Outdoor Controlled Tests}
These tests are similar to the simulated decision-making evaluation but the simulation is replaced by a mock-up test field with traffic signs and obstacles.
This evaluation is also task driven and the vehicle must respect all restrictions and traffic laws.
Their main purpose is to validate the subsystem integration.



\subsection{Autonomous Test and Deployment}
It is impossible to predict every single situation an autonomous car can face.
The occurrence of rare situations is one of the reasons that road field tests are so important.

Initially a vehicle should be tested in real urban roads without implementing the action layer.
The autonomous test step requires human drivers to move around while the autonomous system runs in background without controlling the actuators.
Its main purpose is to use the software logs to find events that could be harmful or that are against the traffic laws.

If a vehicle is able to comply with all these steps it should be safe enough to be deployed in urban roads to start real autonomous tests.

\section{Conclusion}
Evaluating an autonomous car system is a very complex problem that should be done from many scientific perspectives like engineering, computer science, psychology and law.
Although car manufacturers wish to quickly develop self-driving cars their ambitions must not come before safety measures.
We have proposed an evaluation model for autonomous road intelligent agents based in percept sequences.
We have also identified critical aspects as the data dependency from perception models and also the critical points that must be addressed when designing decision making systems.
Since this work focus on perception and decision-making systems it is not meant to become the single procedure to assess an autonomous car performance.
We hope this paper can instigate further discussion about the topic.

\section*{Acknowledgement}

This work was funded by Sao Paulo Research Foundation (FAPESP) project: \#2015/26293-0.

\bibliographystyle{./sty/IEEEtran}
\bibliography{./src/eval}

\end{document}